\documentclass[11pt,a4paper]{article}

\usepackage[utf8]{inputenc}
\usepackage[T1]{fontenc}
\usepackage[english]{babel}
\usepackage{lmodern}

\usepackage[margin=2.5cm]{geometry}
\usepackage{setspace}
\usepackage{titlesec}
\usepackage{enumitem}
\usepackage{parskip}
\setlist[itemize]{label=\textbullet}

\usepackage{booktabs}
\usepackage{array}
\usepackage{longtable}
\usepackage{colortbl}
\usepackage{xcolor}

\usepackage{graphicx}
\usepackage{caption}
\usepackage{float}
\usepackage{tikz}
\usetikzlibrary{arrows.meta, positioning, calc}

\usepackage{amsmath}
\usepackage{amssymb}

\usepackage{url}
\usepackage[colorlinks=true, linkcolor=blue, citecolor=blue, urlcolor=blue]{hyperref}

\graphicspath{{graphes/}{../graphes/}}

\titleformat{\section}{\normalfont\Large\bfseries}{\thesection.}{0.6em}{}
\titleformat{\subsection}{\normalfont\large\bfseries}{\thesubsection}{0.6em}{}

\title{\textbf{Predicting Residential Rents in Dakar Using Machine Learning}}
\author{Amadou Tidiane kassa Diallo \\ Dakar, Senegal \\ \href{mailto:kassadiallo@gmail.com}{kassadiallo@gmail.com}}
\date{}

\begin{document}

\maketitle

\section{Introduction}

Dakar's residential rental market is a particularly relevant field of study for analyzing the mechanisms of rent formation. Despite its economic and social importance, this market remains relatively under-documented from a quantitative standpoint. The Dakar region had 4,004,426 inhabitants in 2023 according to the Agence nationale de la statistique et de la démographie (ANSD). It thus concentrates a significant share of Senegal's population within a relatively small territory. This high demographic concentration is accompanied by strong housing demand and a particularly developed rental market. According to ANSD data, 54.4\% of households in Dakar are renters, compared to 23.3\% nationwide.

In this context, understanding and evaluating rents represents an important issue for households, landlords, real estate professionals, and public authorities. However, Dakar's rental market remains characterized by relatively fragmented information. The prices available on real estate platforms mainly correspond to asking prices rather than actually negotiated prices. Information on completed transactions also remains difficult to access. This situation can create an information asymmetry between the various market participants and make it difficult to objectively assess the rental value of a property.

This issue can be studied through the hedonic pricing theory proposed by Rosen in 1974. According to this approach, the price of a real estate asset implicitly depends on its various characteristics, such as surface area, location, number of bedrooms, and amenities. This theory has for several decades constituted an important reference in the economic analysis of real estate markets and has been widely used in econometric models designed to explain variations in housing prices.

Machine learning has progressively complemented these traditional approaches by making it possible to model nonlinear relationships and complex interactions between housing characteristics. Random forests in particular have shown their value for real estate price prediction. Gradient boosting methods such as XGBoost and LightGBM have subsequently made it possible to achieve high performance on tabular data. Several studies have thus demonstrated the value of these methods for real estate price prediction, particularly in contexts where the relationships between housing characteristics and their prices are difficult to represent with linear models.

The literature devoted to African real estate markets nevertheless remains relatively fragmented. Studies based on the hedonic approach have notably been conducted in Lagos, with the work of Abidoye in 2018, and in Kampala, with that of Irumba in 2015. Approaches based on machine learning have also been applied more recently to various African real estate markets.

As in the short-term rental sector, the 2025 study by Samb devoted to the Greater Banjul region constitutes a particularly relevant reference for the West African context. This study compares several tree-based machine learning models and uses SHAP values to interpret model predictions. This approach provides an interesting methodological framework for analyzing the determinants of real estate prices in a regional context close to that of Dakar.

Despite these advances, Dakar's long-term residential rental market remains little studied using interpretable machine learning methods. To our knowledge, no study has so far proposed a systematic analysis combining rent prediction, model optimization, SHAP interpretability, and uncertainty quantification based on a corpus of Dakar real estate listings.

This article aims to address this gap by developing a complete pipeline for collecting, preparing, modeling, and interpreting real estate data. The objective is not only to obtain a first dataset on rentals in Dakar, but also to identify the main determinants of prices and to assess the uncertainty associated with the predictions.

The main contributions of this study are as follows

\begin{itemize}[leftmargin=1.2em]
\item The construction of an original dataset of 1,507 rental listings in Dakar, obtained through systematic scraping and subjected to a documented cleaning and normalization pipeline
\item The optimization of the XGBoost and LightGBM models through Bayesian search with Optuna, in order to identify the configurations offering the best predictive performance
\item A comparative analysis of feature importance based on XGBoost gain and SHAP values, in order to study the main determinants of rents
\item A quantification of predictive uncertainty via quantile regression, allowing predictions to be complemented with prediction intervals and their calibration to be assessed
\end{itemize}

This approach thus aims to combine predictive performance, interpretability, and uncertainty quantification, while taking into account the specificities and limitations of the data available on Dakar's residential rental market.

\section{Literature Review}

We first briefly present the four families of algorithms evaluated in this study, as well as the two interpretability approaches selected, before detailing the complete methodology in the following section.

\subsection{Linear Regression}

Linear regression estimated by the ordinary least squares method constitutes the reference model of this study. It relies on the assumption of a linear relationship between the explanatory variables and the target variable. This model thus provides a benchmark for evaluating the performance gain brought by approaches capable of modeling nonlinear relationships.

\subsection{Random Forest}

Random Forest is an ensemble learning method based on the construction of a large number of decision trees. Each tree is trained on a bootstrap sample of the data and, at each step, uses a random subset of the explanatory variables. The individual tree predictions are then aggregated by averaging in the case of regression.

This dual source of randomization helps reduce the model's variance and limit the overfitting associated with a single decision tree considered in isolation. Direct interpretation of the results is less straightforward than in the case of linear regression.

\subsection{XGBoost (Extreme Gradient Boosting)}

XGBoost is a gradient boosting method that builds trees sequentially. Each new tree aims to correct the residual errors produced by the previously built trees. Learning is thus based on gradient descent applied to a regularized loss function. When building the trees, the choice of splits is notably based on the gain, which corresponds to the local reduction in the loss function induced by a split.

\subsection{LightGBM}

LightGBM is an implementation of gradient boosting designed to improve computational efficiency and reduce memory usage. Unlike methods that grow trees level by level, LightGBM favors leaf-wise growth, which generally makes it possible to achieve good performance with reduced training time. The model also relies on the discretization of numerical variables using histograms in order to speed up the search for the best splits. It also offers native handling of categorical variables, with training times generally shorter than those of XGBoost for comparable data volumes.

\subsection{Interpretability}

Two complementary approaches are used to interpret the tree-based models: gain-based importance and SHAP values.

Gain-based importance is a measure directly provided by tree-based models. For each variable, it evaluates the average reduction in the loss function associated with the splits in which that variable is involved. The values are then normalized so that their sum corresponds to 100\% across all variables. This measure has the advantage of being fast to compute, but it reflects an average importance at the level of splits rather than a cumulative impact on predictions. Thus, a boolean variable used in a limited number of particularly decisive splits can obtain a high importance, whereas a continuous variable used frequently but with a smaller individual effect may appear less important (see \S4.7).

SHAP (SHapley Additive exPlanations) values, for their part, are based on Shapley values from cooperative game theory. They make it possible to attribute to each variable, for each observation, a specific contribution to the difference between the model's prediction and its reference value. For tree-based models, the TreeSHAP algorithm makes it possible to compute these contributions exactly and efficiently. An essential property of SHAP values is their local additivity. For each observation, the sum of the contributions of all variables corresponds exactly to the gap between the individual prediction and the model's reference prediction. This property distinguishes SHAP values from global importance measures such as gain-based importance, which do not make it possible to directly identify the contribution of each variable to an individual prediction.

\section{Materials and Methods}

\subsection{Study Area}

Dakar is the capital of Senegal. It is located on the Cap Vert peninsula, at the western tip of Senegal. It constitutes the country's main economic and administrative hub and experiences strong pressure on land and rental prices, particularly in coastal areas and the main business districts such as Almadies, Ngor, Ouakam, Point E, and Dakar Plateau.

The corpus covers 24 localities, ranging from upscale residential areas such as Almadies, Ngor, and the Corniche to more working-class or peripheral neighborhoods such as Parcelles Assainies, Grand Yoff, Guédiawaye, and Keur Massar.

\subsection{Dataset}

Data were collected through automated scraping of online real estate listing platforms between January 2024 and June 2025. The collection focuses on residential properties for rent in the Dakar region.

The initial collection comprises 1,654 listings and 36 variables. A seven-step preprocessing procedure was then applied in order to obtain a homogeneous dataset.

\begin{itemize}[leftmargin=1.2em]
\item Filtering listings by locality
\item Removing listings without a valid rental price
\item Restricting to properties intended for residential rental
\item Normalizing housing types into six categories: F2, F3, F4, F5, F6, and Studio
\item Excluding short-term rentals
\item Standardizing more than 130 locality labels into 24 geographic categories
\item Handling outliers, missing values, and redundant variables
\end{itemize}

\subsection{Feature Engineering}

Four derived variables were created in order to enrich the information available in the raw data.

\begin{itemize}[leftmargin=1.2em]
\item \textbf{luxury\_score} is a synthetic score based on seven attributes associated with the upscale character of the property. It takes into account the swimming pool, sea view, penthouse, standing, elevator, garage, and air conditioning.
\item \textbf{nbre\_chambre\textsuperscript{2}} is a quadratic term used to capture a possible nonlinear relationship between the number of bedrooms and the rent level.
\item \textbf{ratio\_sdb\_ch} corresponds to the ratio between the number of bathrooms and the number of bedrooms. It serves as a proxy variable for the property's comfort level.
\item \textbf{kw\_score} is a weighted score built from 18 keywords associated with the premium positioning of properties. Each term is given a weight according to its indicative power. The terms "corniche" and "haut standing" are for example weighted at 3, while "piscine" receives a weight of 2 and "meublé" a weight of 1.
\end{itemize}

Following preprocessing and feature engineering, the final dataset comprises \textbf{1,507 listings and 33 variables}. The structure and nature of the main variables retained are presented in the following table.

\begin{table}[H]
\centering
\small
\begin{tabular}{|l|p{8.5cm}|l|}
\hline
\textbf{Variable} & \textbf{Description} & \textbf{Type} \\
\hline
price & Monthly rent requested (FCFA) --- target variable & int \\
\hline
location & Dakar locality (24 canonical values) & categorical \\
\hline
type\_appartement & Normalized type: F2, F3, F4, F5, F6, Studio & categorical \\
\hline
surface & Living area (m\textsuperscript{2}) & float \\
\hline
nbre\_chambre & Number of bedrooms & int \\
\hline
nbre\_salle\_de\_bain & Number of bathrooms & int \\
\hline
etage & Floor (0 = ground floor) & int \\
\hline
luxury\_score & Weighted sum of 7 premium attributes & int \\
\hline
kw\_score & Premium keyword score extracted from the description (derived) & int \\
\hline
nbre\_chambre\textsuperscript{2} & the square of the number of bedrooms & int \\
\hline
23 boolean columns & Amenities: elevator, swimming pool, parking, air conditioning, sea view, standing, furnished, etc. & bool \\
\hline
\end{tabular}
\caption{Description of dataset variables}
\label{tab:variables}
\end{table}

\subsection{Methodological Pipeline}

The methodological approach comprises five main steps: data preparation, feature engineering, preprocessor setup, model training and optimization, and finally their evaluation and interpretation. The whole is structured within a coherent pipeline presented in the figure below.

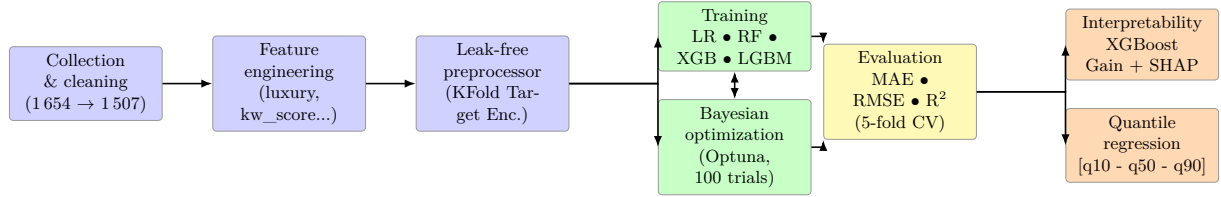
\begin{figure}[H]
\centering
\resizebox{\textwidth}{!}{%
\begin{tikzpicture}[
    font=\footnotesize,
    box/.style={rectangle, rounded corners=2pt, draw=black!40, minimum width=2.6cm, minimum height=1.2cm, align=center, text width=2.5cm},
    blue box/.style={box, fill=blue!18},
    green box/.style={box, fill=green!22},
    yellow box/.style={box, fill=yellow!35},
    orange box/.style={box, fill=orange!30},
    arr/.style={-{Latex[length=2mm]}, thick}
]
\node[blue box] (collecte) {Collection \& cleaning\\(1\,654 $\rightarrow$ 1\,507)};
\node[blue box, right=0.9cm of collecte] (feat) {Feature\\engineering\\(luxury, kw\_score...)};
\node[blue box, right=0.9cm of feat] (prepro) {Leak-free\\preprocessor\\(KFold Target Enc.)};

\node[green box, right=1.6cm of prepro, yshift=0.85cm] (train) {Training\\LR $\bullet$ RF $\bullet$ XGB $\bullet$ LGBM};
\node[green box, below=0.5cm of train] (optim) {Bayesian\\optimization\\(Optuna, 100 trials)};
\coordinate (col4mid) at ($(train)!0.5!(optim)$);

\node[yellow box, right=1.6cm of col4mid] (eval) {Evaluation\\MAE $\bullet$ RMSE $\bullet$ R\textsuperscript{2}\\(5-fold CV)};

\node[orange box, right=1.6cm of eval, yshift=0.85cm] (interp) {Interpretability\\XGBoost Gain + SHAP};
\node[orange box, below=0.5cm of interp] (quant) {Quantile regression\\{}[q10 - q50 - q90]};

\draw[arr] (collecte) -- (feat);
\draw[arr] (feat) -- (prepro);
\draw[arr] (prepro.east) -| (train.west);
\draw[arr] (prepro.east) -| (optim.west);
\draw[arr, {Latex[length=2mm]}-{Latex[length=2mm]}] (train) -- (optim);
\draw[arr] (train.east) -| (eval.north west);
\draw[arr] (optim.east) -| (eval.south west);
\draw[arr] (eval.east) -| (interp.west);
\draw[arr] (eval.east) -| (quant.west);
\end{tikzpicture}%
}
\caption{pipeline diagram}
\label{fig:pipeline}
\end{figure}

\subsection{Leak-Free KFold Target Encoding}

The target variable exhibits strong right skewness, with a skewness coefficient of 1.55. A logarithmic transformation is therefore applied according to the relation

\begin{equation}
y = \log(1 + \text{price})
\end{equation}

Predictions are then converted back to the original scale using the inverse transformation

\begin{equation}
\widehat{\text{price}} = \exp(\hat{y}) - 1
\end{equation}

The 33 explanatory variables are divided into four blocks processed within a scikit-learn \texttt{ColumnTransformer}.

\begin{itemize}[leftmargin=1.2em]
\item \textbf{Numerical block} composed of 8 variables. Missing values are replaced with the median, and the variables are then standardized using \texttt{StandardScaler}
\item \textbf{Categorical block} corresponding to the \texttt{type\_appartement} variable. Missing values are replaced with the most frequent category, and the variable is then transformed via One-Hot Encoding
\item \textbf{Location block} processed via KFold Target Encoding in order to incorporate neighborhood-related information while avoiding data leakage
\item \textbf{Boolean block} grouping the 23 amenities. Missing values are replaced with the most frequent category
\end{itemize}

A naive target encoding consists of computing, for each locality, a price statistic based on the entire set of observations before splitting the data into training and test sets. This practice introduces data leakage and can lead to an overestimation of performance.

To avoid this bias, a \texttt{KFoldTargetEncoder}, implemented as a scikit-learn transformer, was used. During training, the encoding is computed solely from the observations available in the training set

\begin{equation}
\mathrm{enc}(\text{neighborhood}_k) = \operatorname{median}\{y_i : \text{neighborhood}_i = k,\ i \in \text{train}\}
\end{equation}

When a locality is absent from the training set and appears in the test data, the overall median of the training set is used. This procedure ensures a strict separation between the training and test data and thus limits the risk of information leakage.

\subsection{Models and Hyperparameter Optimization}

Five modeling approaches were compared. Linear regression serves as the reference model. It is compared to a Random Forest configured with 300 trees, a maximum depth of 20, and a minimum of 2 observations per leaf. A baseline XGBoost is also used with 1,000 estimators, a learning rate of 0.03, and a maximum depth of 6. Finally, optimized versions of XGBoost and LightGBM are trained.

The hyperparameters of the optimized models are determined through Bayesian optimization with Optuna using the Tree-structured Parzen Estimator algorithm. The search space comprises 8 to 9 hyperparameters, covering notably the number of trees, the learning rate, the maximum depth, the observation and feature subsampling rates, as well as the L1 and L2 regularizations and the minimum weight associated with a leaf.

\subsection{Evaluation Metrics}

Performance is evaluated using three main metrics: MAE, RMSE, and the coefficient of determination R\textsuperscript{2}. The metrics are computed after inverse transformation of the predictions in order to preserve the original price scale in FCFA.

\begin{equation}
MAE = \frac{1}{n}\sum |\hat{y}_i-y_i|
\end{equation}

\begin{equation}
RMSE = \sqrt{\frac{1}{n}\sum(\hat{y}_i-y_i)^2}
\end{equation}

\begin{equation}
R^2 = 1-\frac{\sum(\hat{y}_i-y_i)^2}{\sum(y_i-\bar{y})^2}
\end{equation}

where $n$ represents the number of observations, $y_i$ the actual value, $\hat{y}_i$ the predicted value, and $\bar{y}$ the mean of the observed values.

The main protocol relies on an 80\%/20\% split of the data for training and testing. A 5-fold cross-validation complements this evaluation in order to examine the robustness and stability of the performance obtained, as presented in Section 4.6.

\subsection{Model Interpretability}

Two complementary approaches are applied to the optimized XGBoost model, retained as the best model. The first relies on the native gain-based importance, computed from the average reduction in the loss function brought about by the different variables in the trees. The second uses SHAP values computed on the test set.

SHAP values allow for a global interpretation based on the mean of the absolute values per variable, as well as a local interpretation through the decomposition of each variable's contributions to an individual prediction. The comparison of these two approaches is presented in Section 4.7.

\subsection{Quantile Regression}

Three additional XGBoost models are trained with the \texttt{reg:quantileerror} objective for the 0.10, 0.50, and 0.90 quantiles. The optimal hyperparameters identified in Section 3.6 are retained.

The interval between q10 and q90 targets a nominal coverage of 80\% on the test set. This approach makes it possible to associate an estimate of predictive uncertainty with each prediction.

\subsection{Software Environment}

The entire pipeline is developed in Python 3.12.3 with pandas 3.0.3, numpy 2.4.6, scikit-learn 1.9.0, XGBoost 3.3.0, LightGBM 4.6.0, Optuna 4.9.0, and SHAP 0.52.0 on a Linux environment.

The complete code, as well as the Jupyter notebook documenting the various steps of the pipeline, are available upon request from the author.

\section{Results and Analysis}

\subsection{Dataset Overview}

The final dataset comprises 1,507 real estate listings spread across 24 localities. The median monthly rent is 700,000 FCFA, compared to a mean of 950,411 FCFA, with a standard deviation of 796,534 FCFA. The median living area is 189~m\textsuperscript{2}, and the median number of bedrooms is 3.

The majority of properties are of type F4, accounting for 53.4\% of listings, followed by F3 at 23.6\%.

The geographic distribution remains concentrated in a few neighborhoods. The most represented are Almadies with 19.5\% of listings, followed by Ouakam with 11.9\%, followed by Mermoz with 10.4\%, followed by Point E with 7.2\%, and Fann with 6.6\%. These five neighborhoods alone account for 55.6\% of the corpus.

\begin{figure}[H]
\centering
\includegraphics[width=0.85\textwidth]{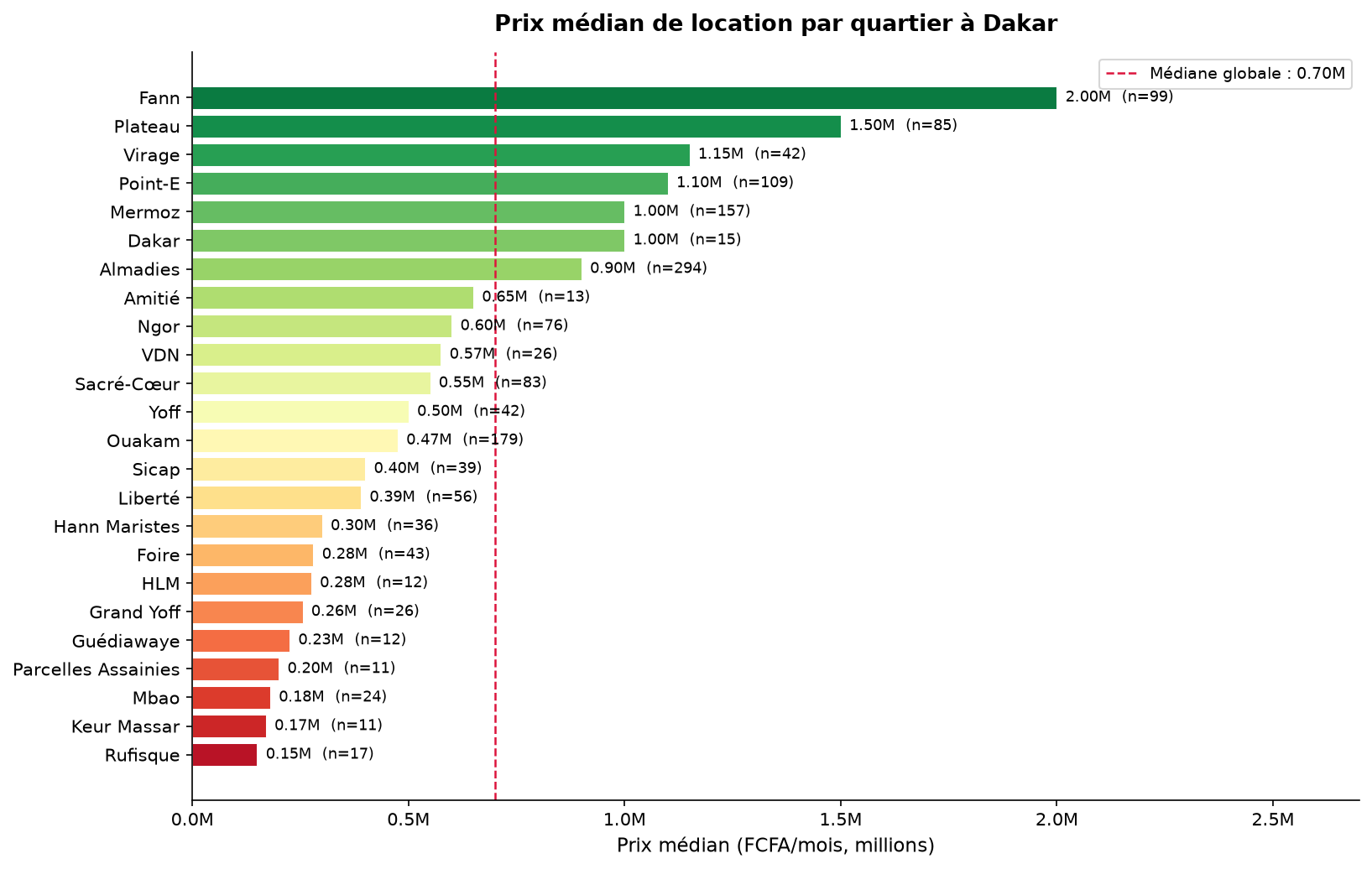}
\caption{median rental price by locality (Dakar neighborhood)}
\label{fig:prix_quartier}
\end{figure}

This geographic concentration represents an imbalance to be taken into account when interpreting the results, particularly when analyzing model performance by neighborhood, as presented in Section 5.4.

\subsection{Price Distribution}

The distribution of rents exhibits strong right skewness, with a skewness coefficient of 1.55. The majority of listings fall between 300,000 and 1,300,000 FCFA per month, while a long tail comprises upscale properties whose rents can exceed 2,000,000 FCFA.

In order to limit the influence of extreme values, a logarithmic transformation of the target variable according to the relation $\log(1 + \text{price})$ was applied before training the models. This transformation reduces the dispersion of high values and produces a more balanced distribution, as shown in the figure below.

\begin{figure}[H]
\centering
\includegraphics[width=0.95\textwidth]{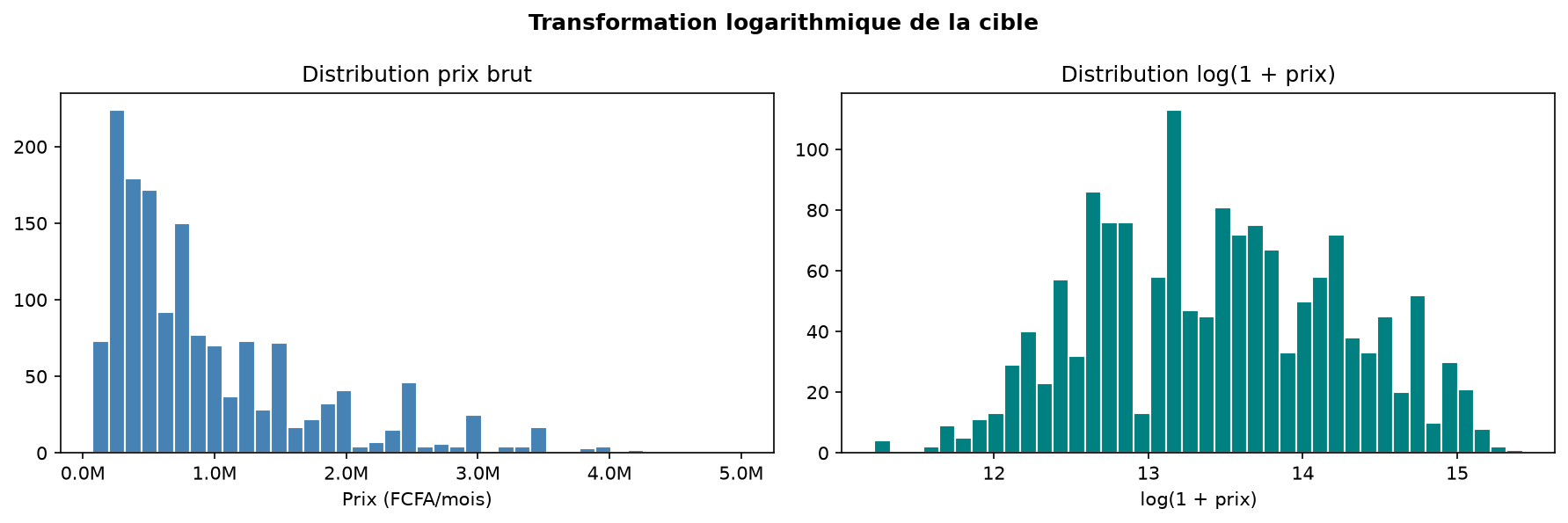}
\caption{logarithmic transformation of price}
\label{fig:log_transform}
\end{figure}

\subsection{Spatial Structure of Prices and Correlations}

The figure below presents the correlation matrix between the main numerical variables. The strongest correlations with price concern the number of bedrooms, the surface area, and the \texttt{luxury\_score}. These results are consistent with the hedonic approach, according to which property characteristics contribute to rent formation.

However, these relationships, taken individually, do not fully explain the variations observed in rents. This limitation justifies the use of models capable of accounting for nonlinear relationships and interactions between the different characteristics of properties.

\begin{figure}[H]
\centering
\includegraphics[width=0.7\textwidth]{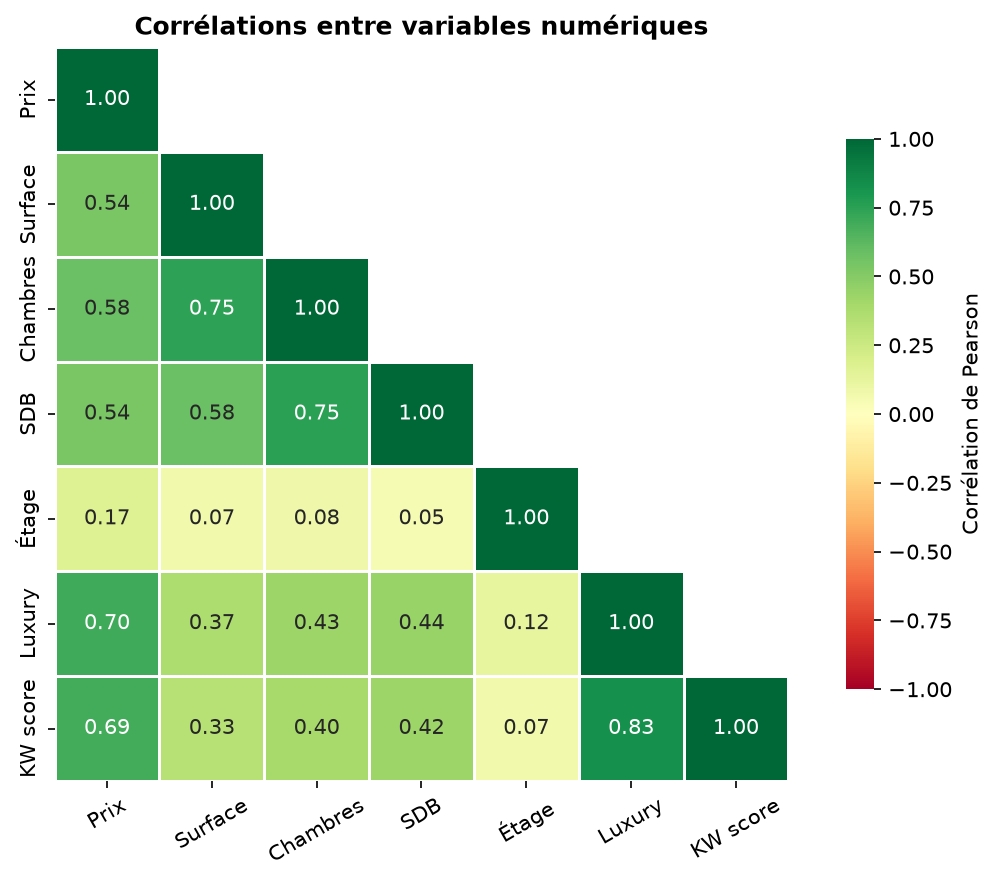}
\caption{correlation matrix}
\label{fig:correlation}
\end{figure}

\subsection{Comparative Model Evaluation}

The table below gathers the performance of the five models on the test set, evaluated after inverse transformation to the FCFA scale.

\begin{table}[H]
\centering
\begin{tabular}{|l|r|r|r|}
\hline
\textbf{Model} & \textbf{MAE (FCFA)} & \textbf{RMSE (FCFA)} & \textbf{R\textsuperscript{2}} \\
\hline
Linear Regression       & 257\,147 & 470\,262 & 0.679 \\
\hline
Random Forest          & 222\,201 & 363\,525 & 0.808 \\
\hline
XGBoost baseline        & 221\,405 & 350\,685 & 0.821 \\
\hline
XGBoost tuned (Optuna)   & 210\,902 & 324\,195 & 0.847 \\
\hline
LightGBM tuned (Optuna)  & 215\,259 & 329\,978 & 0.842 \\
\hline
\end{tabular}
\caption{Comparative performance on the test set}
\label{tab:perf}
\end{table}

The XGBoost model optimized with Optuna achieves the best performance, with an R\textsuperscript{2} of 0.847, closely followed by the optimized LightGBM with an R\textsuperscript{2} of 0.842. The gap of 16.8 R\textsuperscript{2} points between linear regression, which reaches 0.679, and XGBoost highlights the importance of nonlinear relationships in determining rents in Dakar. These results show the value of machine learning models in capturing the interactions and complex effects between the various characteristics of properties.

\subsection{Prediction Accuracy}

\begin{figure}[H]
\centering
\includegraphics[width=0.85\textwidth]{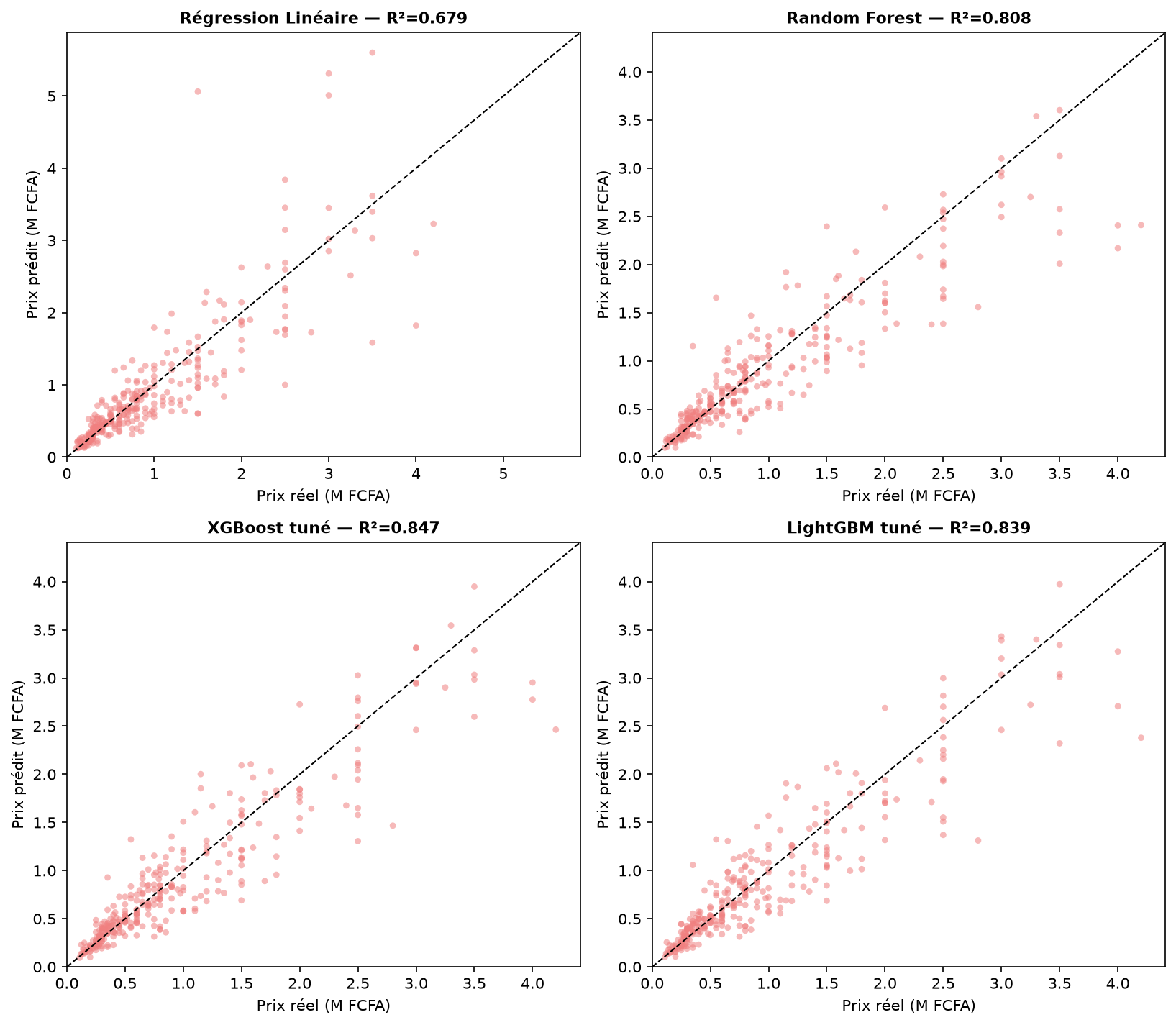}
\caption{Actual vs. predicted price on the test set}
\label{fig:reel_vs_predit}
\end{figure}

Linear regression shows greater dispersion around the diagonal, particularly for higher rents. The optimized XGBoost and LightGBM models produce a tighter cluster of points. A significant residual dispersion is nevertheless still observed for certain upscale properties whose rent exceeds 2,000,000 FCFA.

\subsection{Cross-Validation and Robustness}

A 5-fold cross-validation was carried out in order to assess the stability of performance beyond the single 80/20 split. Table 3 presents the mean R\textsuperscript{2} and its standard deviation, computed on the $\log(1 + \text{price})$ scale used for training the models.

\begin{table}[H]
\centering
\begin{tabular}{|l|r|r|r|}
\hline
\textbf{Model} & \textbf{Mean R\textsuperscript{2} (log)} & \textbf{R\textsuperscript{2} Std. Dev.} & \textbf{Mean RMSE (log)} \\
\hline
Linear Regression & 0.814 & 0.013 & 0.350 \\
\hline
Random Forest        & 0.832 & 0.021 & 0.331 \\
\hline
XGBoost baseline      & 0.839 & 0.015 & 0.325 \\
\hline
XGBoost tuned          & 0.852 & 0.011 & 0.312 \\
\hline
LightGBM tuned         & 0.851 & 0.013 & 0.312 \\
\hline
\end{tabular}
\caption{5-fold cross-validation}
\label{tab:cv5}
\end{table}

The results obtained on the logarithmic scale during cross-validation confirm the performance observed on the single split. The optimized XGBoost and LightGBM models achieve the best results, with a mean R\textsuperscript{2} of approximately 0.852.

In order to more precisely assess the robustness of the retained model, optimized XGBoost, a second 5-fold cross-validation was carried out directly on the original price scale in FCFA, after inverse transformation. This approach makes it possible to express performance in directly interpretable monetary units and to obtain a more concrete measure of prediction error.

\begin{figure}[H]
\centering
\includegraphics[width=0.95\textwidth]{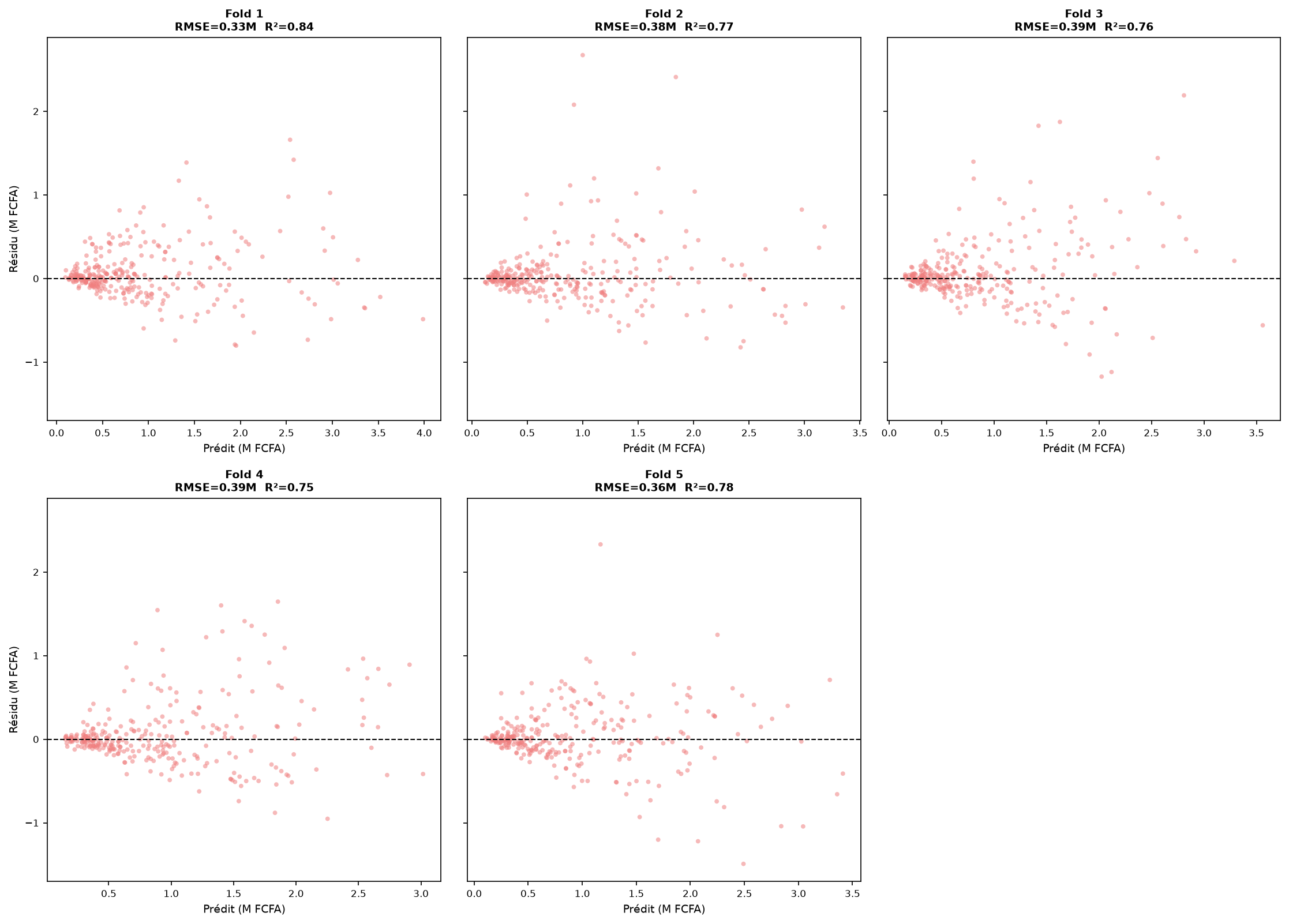}
\caption{Distribution of residuals by fold}
\label{fig:residus_fold}
\end{figure}

This check reveals a gap between the performance obtained via cross-validation and that presented in Section 4.4. The mean R\textsuperscript{2} reaches \textbf{0.782}, with a standard deviation of \textbf{0.034} and a range between \textbf{0.750 and 0.848}, compared to 0.847 for the single 80/20 split. The mean RMSE stands at \textbf{369,744 FCFA}, with a standard deviation of \textbf{26,230 FCFA}, compared to 324,195 FCFA for the single split.

The single split should be interpreted with caution and complemented by cross-validation in order to assess the robustness and stability of the model.

\subsection{Feature Importance According to XGBoost Gain}

The analysis of feature importance according to XGBoost's native metric, based on the average gain per split, highlights several variables strongly associated with the model's predictions. The main ones are \texttt{luxury\_score} with 23.7\% of the total gain, followed by \texttt{nbre\_chambre} with 14.4\%, \texttt{nbre\_chambre\_sq} with 12.3\%, \texttt{vue\_mer} with 7.7\%, \texttt{kw\_score} with 7.3\%, \texttt{location\_encoded} with 3.4\%, and \texttt{parking} with 2.9\%.

The 23 binary variables corresponding to amenities collectively account for 28\% of the total gain. This substantial contribution should nevertheless be interpreted with caution, as some variables may be correlated or provide similar information to the model.

The comparison with SHAP values presented in the following section makes it possible to complement this analysis and to better assess the actual influence of the different variables on the predictions.

\begin{figure}[H]
\centering
\includegraphics[width=0.85\textwidth]{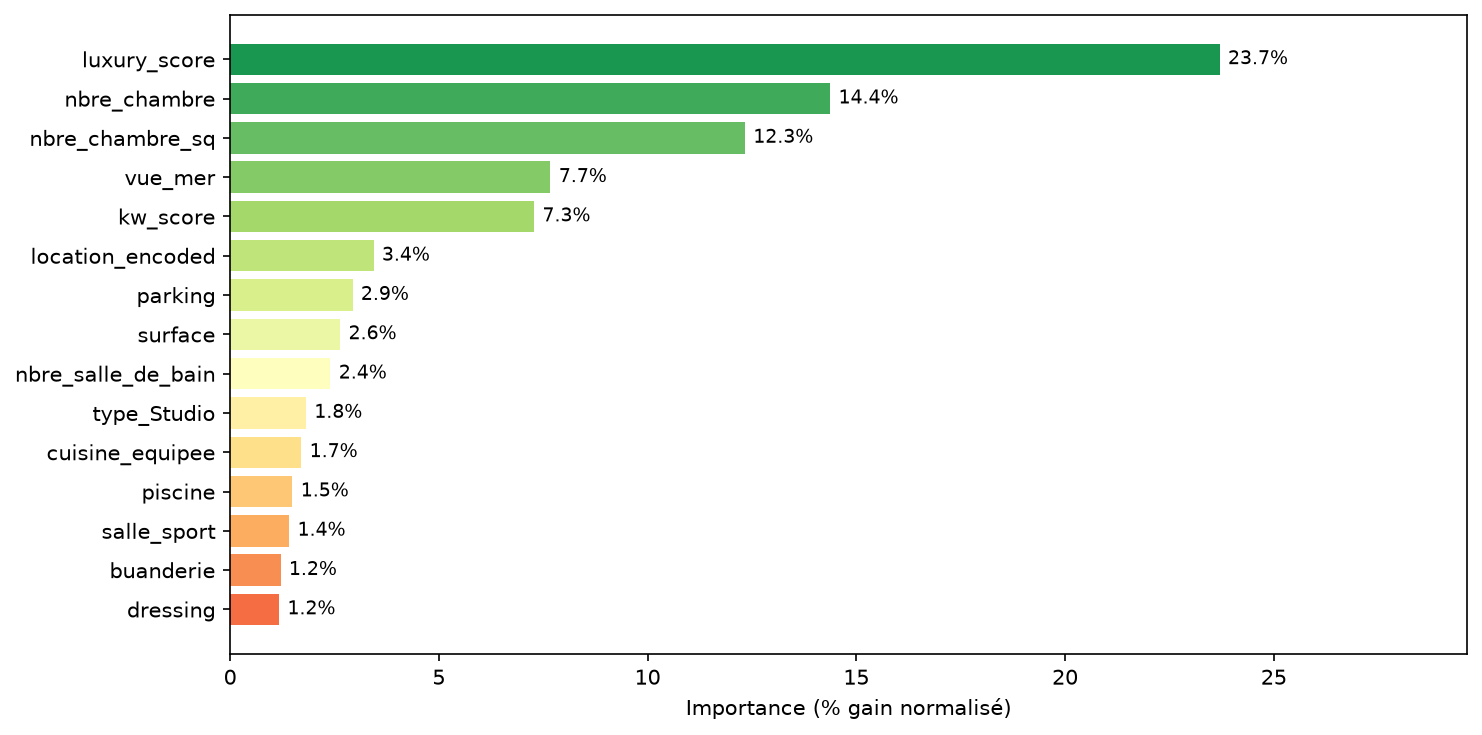}
\caption{Feature importance by gain}
\label{fig:gain_importance}
\end{figure}

\subsection{Interpretability via SHAP Values}

In order to more precisely analyze the influence of variables on the model's predictions, SHAP values were computed on the test set.

The model's base value, corresponding to the mean predicted log price, is \textbf{13.44}, or approximately \textbf{686,870 FCFA} after conversion back to the price scale.

As shown in Figure 8, the ranking based on the mean of the absolute SHAP values highlights the most influential variables. The six main ones are \texttt{luxury\_score} with 0.163, followed by \texttt{location\_encoded} with 0.155, \texttt{kw\_score} with 0.132, \texttt{surface} with 0.107, \texttt{nbre\_chambre} with 0.075, and \texttt{vue\_mer} with 0.072.

\begin{figure}[H]
\centering
\includegraphics[width=0.85\textwidth]{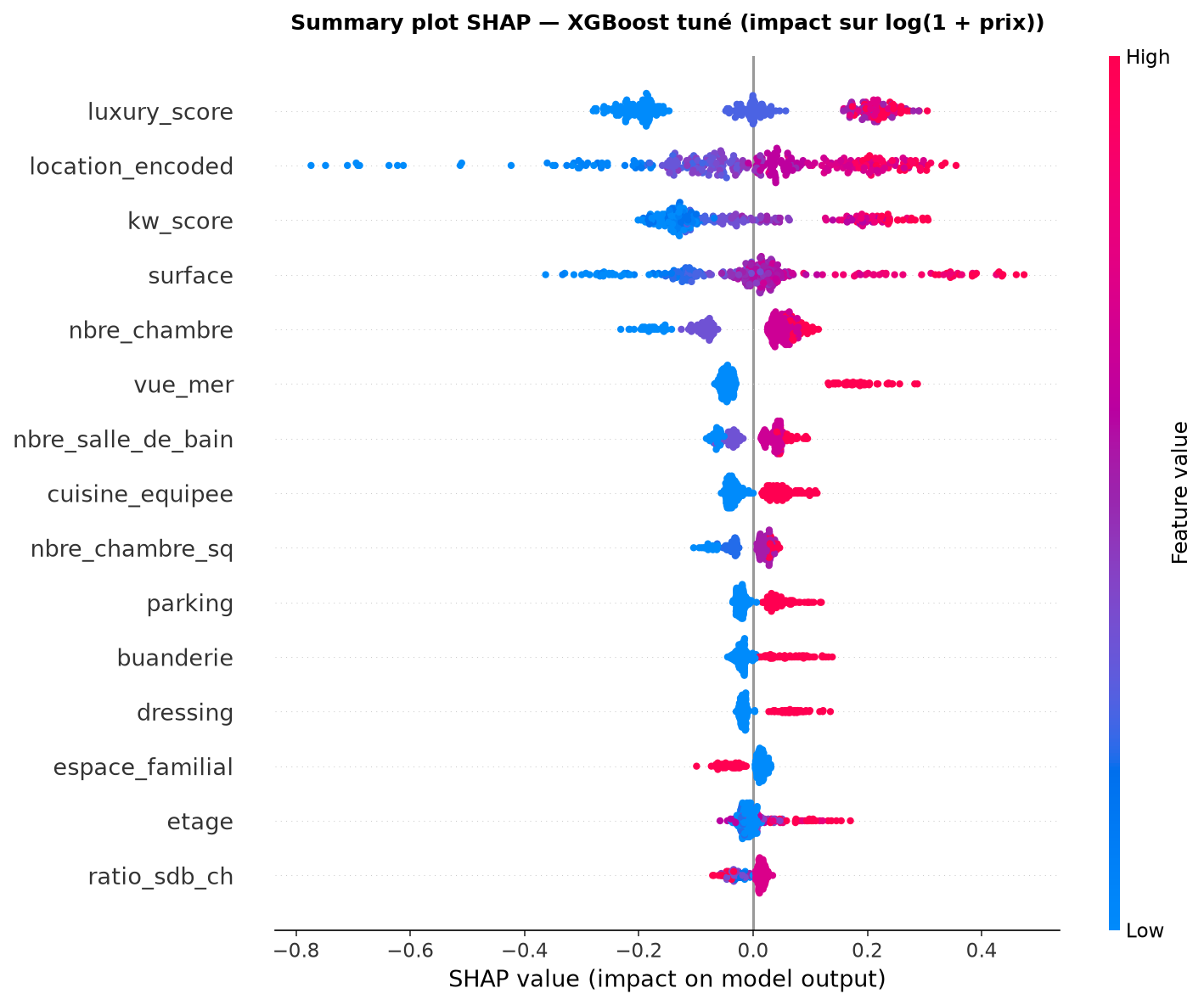}
\caption{SHAP summary plot}
\label{fig:shap_summary}
\end{figure}

\begin{figure}[H]
\centering
\includegraphics[width=0.85\textwidth]{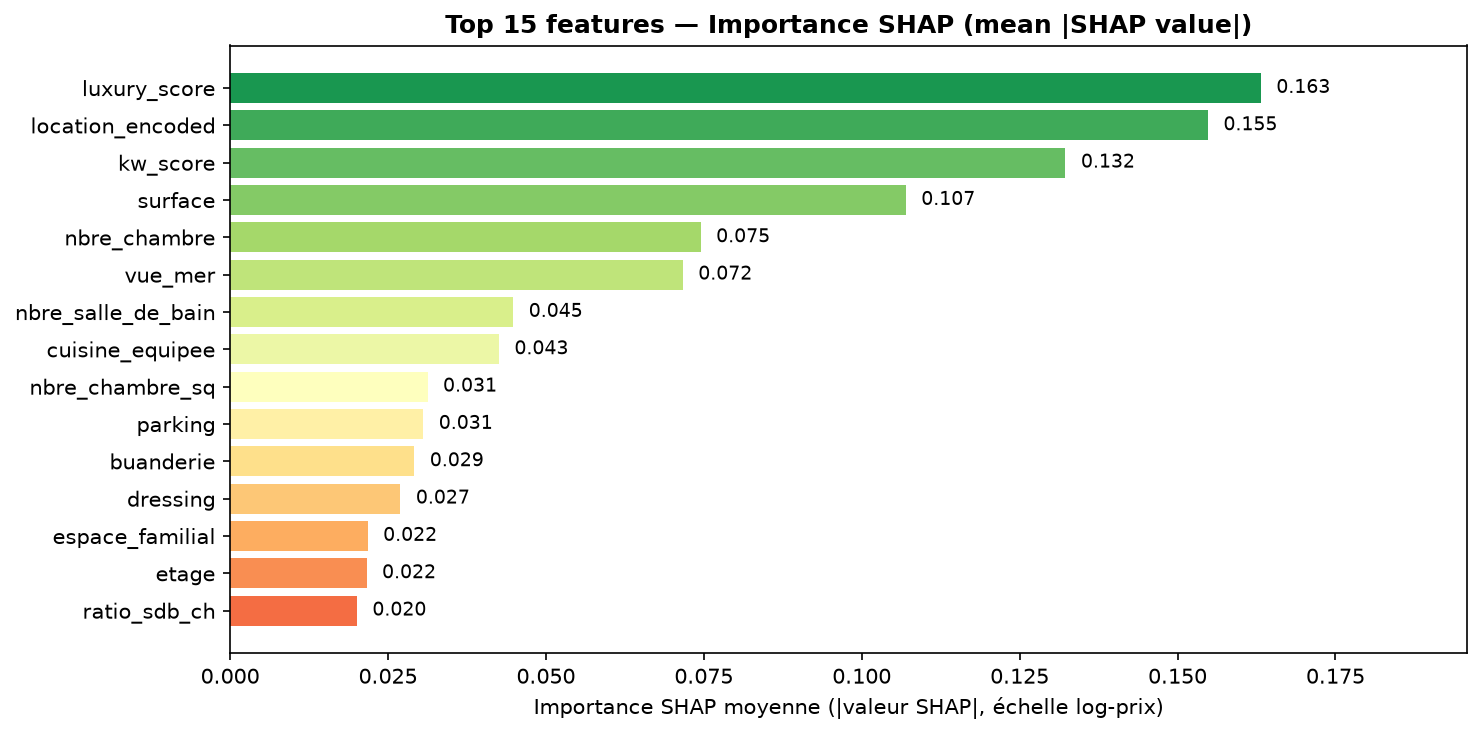}
\caption{Mean SHAP importance (\textbar SHAP value\textbar)}
\label{fig:shap_importance}
\end{figure}

Unlike XGBoost's native importance presented in Section 4.7, SHAP values make it possible to assess the contribution of each variable to individual predictions. They thus offer a complementary interpretation of feature importance and make it possible to better understand their actual influence on the model's predictions.

The scatter plot of the summary plot (Figure 8) confirms, for the top six features, a monotonically increasing relationship between the feature value (color) and its SHAP impact (horizontal position). Pearson correlation coefficients between feature value and the associated SHAP value exceed 0.90 for \texttt{location\_encoded}, \texttt{luxury\_score}, \texttt{kw\_score}, \texttt{surface}, \texttt{nbre\_chambre}, and \texttt{vue\_mer} (Table~4).

\begin{table}[H]
\centering
\begin{tabular}{|l|r|}
\hline
\textbf{Feature} & \textbf{Correlation (value, SHAP)} \\
\hline
location\_encoded & 0.932 \\
\hline
luxury\_score      & 0.897 \\
\hline
kw\_score          & 0.952 \\
\hline
surface            & 0.920 \\
\hline
nbre\_chambre      & 0.910 \\
\hline
vue\_mer           & 0.993 \\
\hline
\end{tabular}
\caption{Correlation between feature value and its associated SHAP value}
\label{tab:shap_corr}
\end{table}

\subsubsection*{Divergence Between Gain and SHAP Values: The Case of Location}

The interpretability analysis reveals a significant difference in the ranking of \texttt{location\_encoded} depending on the measure used. With XGBoost gain-based importance, this variable occupies the 6th position with 3.4\% of the total gain. With the mean of the absolute SHAP values, it ranks 2nd, behind \texttt{luxury\_score}.

This difference does not reflect model instability but rather results from the calculation principles specific to the two measures. XGBoost gain evaluates the average reduction in the loss function brought about by a split, and thus does not directly measure a variable's overall impact on the full set of predictions. Since \texttt{location\_encoded} is a continuous variable resulting from a KFold encoding of location, it can appear in many splits at different thresholds. Each split may then bring a relatively small improvement. Conversely, a binary variable such as \texttt{vue\_mer} offers fewer splitting possibilities. When it is used for a split, that split can be highly discriminative and produce a high average gain despite less frequent use.

SHAP values take a different approach by measuring each variable's contribution to individual predictions and then aggregating these contributions across all observations. They thus make it possible to better represent the actual influence of variables on the model's predictions. The resulting ranking confirms that location constitutes one of the main determinants of rents in Dakar. This result is consistent with the hedonic approach and with the results presented in Figure 4 and the correlation Table 2 in Section 4.3.

This difference also holds methodological interest. When location is represented by a single continuous variable derived from target encoding, its importance may be underestimated by the gain measure in tree-based models. The use of SHAP values as a complement to XGBoost's native importance therefore appears relevant for obtaining a more complete interpretation of the variables.

\subsubsection*{Local Explanation via Waterfall Decomposition}

SHAP values make it possible to explain each individual prediction thanks to their additivity property. A prediction can thus be decomposed starting from the model's base value and then completed by the contribution of each variable.

For a listing whose predicted price deviates from the average market level, this decomposition makes it possible to identify the characteristics that explain this gap. It can notably highlight the influence of location, surface area, number of bedrooms, and amenities.

This approach provides a detailed and verifiable interpretation of individual predictions. An example is presented in the accompanying notebook in the form of a waterfall chart.

\begin{figure}[H]
\centering
\includegraphics[width=0.75\textwidth]{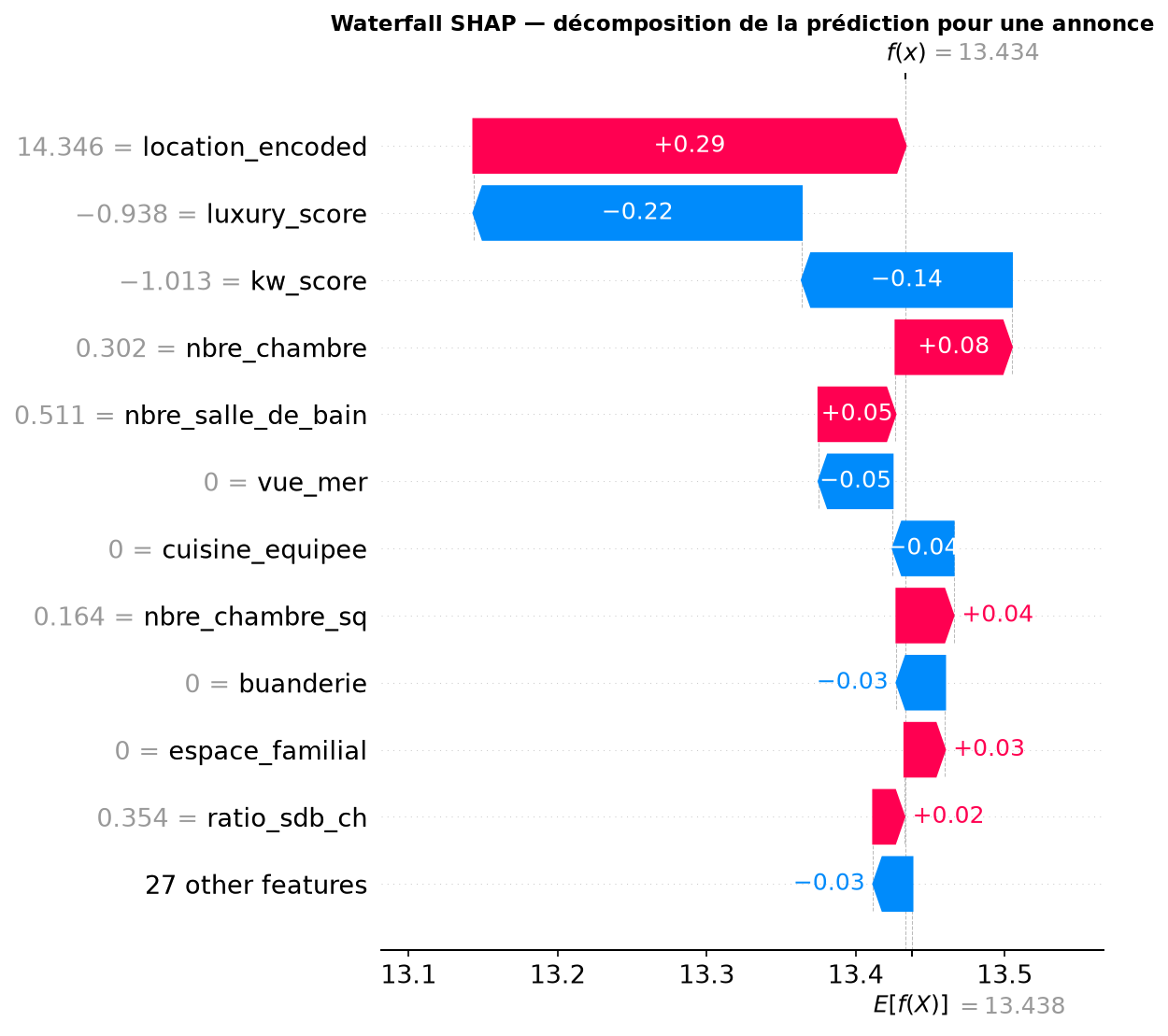}
\caption*{\textit{SHAP Waterfall --- decomposition of the prediction for a listing}}
\label{fig:waterfall}
\end{figure}

\subsection{Prediction Intervals via Quantile Regression}

In order to quantify the uncertainty associated with the predictions, three additional XGBoost models corresponding to the q10, q50, and q90 quantiles were trained with the \texttt{reg:quantileerror} objective and the optimal hyperparameters defined in Section 3.6.

The q10 and q90 quantiles define a prediction interval targeting a nominal coverage of 80\% of observations. The q50 quantile corresponds to the median rent estimate.

This method thus makes it possible to complement the point prediction with an uncertainty estimate and to assess the possible dispersion of rents around the predicted value.

\begin{figure}[H]
\centering
\includegraphics[width=0.85\textwidth]{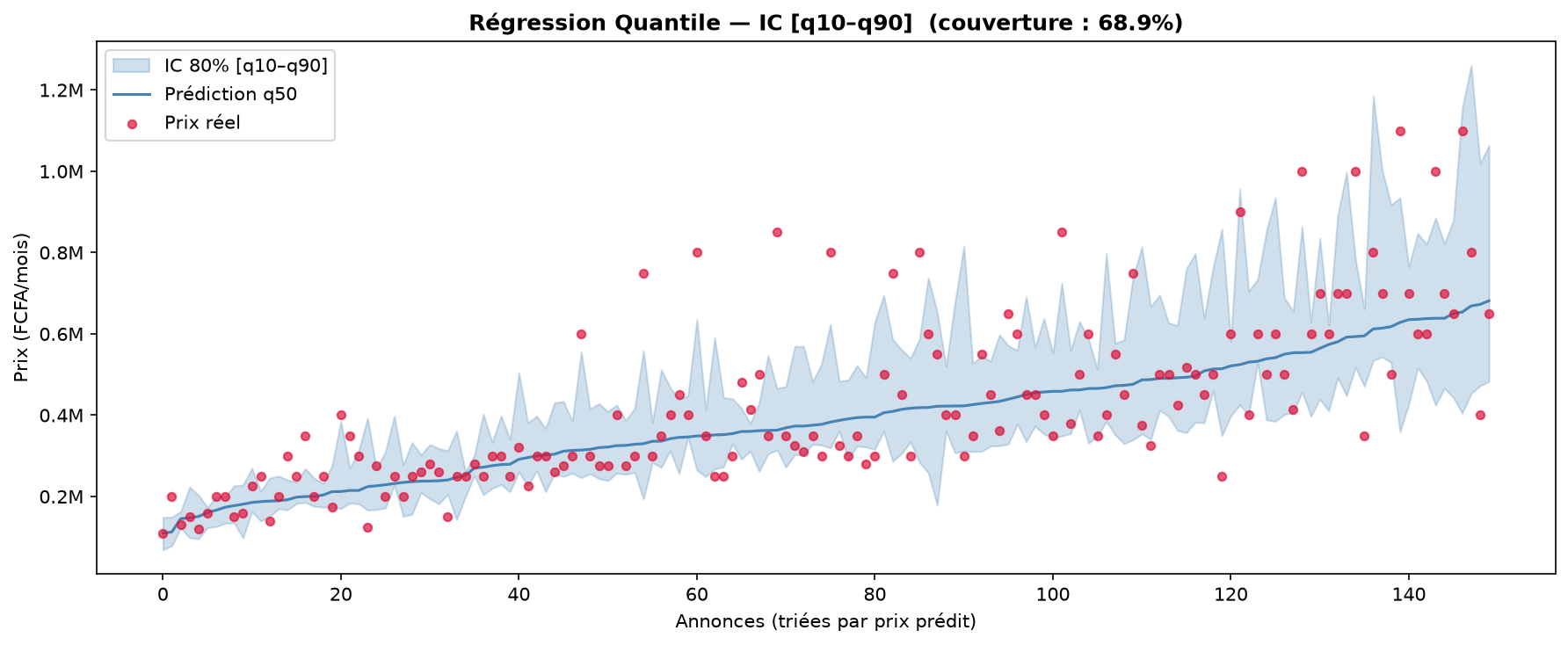}
\caption*{\textit{Quantile Regression --- CI [q10--q90]}}
\label{fig:quantile}
\end{figure}

The \textbf{empirical coverage rate} obtained on the test set is \textbf{64.2\%}, which remains well below the \textbf{targeted nominal coverage of 80\%}. The \textbf{median width of the prediction intervals} reaches \textbf{376,376 FCFA}.

This undercoverage indicates that the intervals produced by the quantile models are not yet properly calibrated. For a significant proportion of listings, the lower bound q10 is too high, as is the upper bound q90, which is too low. The resulting intervals are thus too narrow relative to the actual variability of rents.

\section{Discussion}

\subsection{Interpretation of Performance}

The R\textsuperscript{2} of 0.847 obtained with the optimized XGBoost model on the single split shows that the characteristics extracted from real estate platforms contain significant information about rent variation in Dakar. This performance remains close to that obtained in other real estate contexts, while being slightly below certain benchmark results.

Truong et al. (2020) notably obtain an R\textsuperscript{2} of 0.88 on an Australian corpus of more than 20,000 transactions. Their study benefits from detailed geospatial variables such as distance to schools, transportation, and green spaces. This information is not available in our corpus. The gap of 3 to 4 R\textsuperscript{2} points can therefore be partly explained by the smaller size of our sample, with 1,507 observations, and by the absence of precise geographic coordinates.

Despite these limitations, the results show that the variables available in real estate listings make it possible to capture a significant share of the rent variation observed in Dakar.

\subsection{Gain and SHAP: Methodological Implications}

The difference observed between XGBoost gain-based importance and that obtained with SHAP values constitutes an important methodological finding of this study. The ranking of \texttt{location\_encoded} varies significantly depending on the measure used. This variable occupies a relatively low position according to gain, whereas it ranks among the most influential variables according to SHAP.

This divergence shows that the choice of interpretability method can significantly alter the analysis of real estate price determinants. In our case, it notably changes the relative importance attributed to location compared to the property's characteristics.

For hedonic studies using ensemble models and target encoding, these results demonstrate the value of not relying on a single importance measure. Combining XGBoost gain and SHAP values allows for a more comprehensive analysis and limits the risk of conclusions based on a single metric.

\subsection{Selection Bias and Price Negotiability}

The corpus likely presents an overrepresentation of upscale properties and listings marketed on online real estate platforms. Part of Dakar's rental market, particularly properties offered through informal networks or without online publication, remains absent from the data.

The results must therefore be interpreted within the limits of the observed distribution and cannot be directly generalized to the entire Dakar rental market.

Another limitation concerns the nature of the prices used. The data correspond to \textbf{asking prices} rather than \textbf{actually negotiated prices}. Since negotiation between landlords and tenants is common, the rents ultimately agreed upon may differ from the amounts displayed.

This difference can introduce an upward bias in the data used to train the model. The predictions should thus be interpreted as estimates of the asking price level rather than as a direct measure of the rent actually paid.

\subsection{Robustness and Limitations of the Corpus}

The concentration of observations in certain Dakar localities also constitutes a limitation for interpreting performance across geographic areas. The most represented neighborhoods have a larger volume of observations, while less represented areas in the corpus may be less well represented by the model.

This unbalanced distribution may limit the model's ability to generalize its predictions in neighborhoods with few observations. A broader and more geographically balanced data collection would improve the representativeness of the corpus and allow for a more precise assessment of performance differences between neighborhoods.

\subsection{Imperfect Calibration of Prediction Intervals}

The empirical coverage rate of 64.2\% obtained in Section 4.9 remains well below the nominal coverage of 80\%. This result constitutes an important limitation for the use of the prediction intervals.

Several factors may explain this undercoverage. The size of the test set, with 302 observations, first limits the precision of the coverage estimate. XGBoost's \texttt{reg:quantileerror} objective can also be sensitive to hyperparameter choices.

The heterogeneity of Dakar's rental market can also complicate the estimation of extreme quantiles. Properties with similar characteristics may indeed display very different rent levels depending on their standing or environment.

The resulting intervals should therefore not be considered perfectly calibrated for operational use. Future work could notably focus on hyperparameter optimization specific to each quantile and on post-hoc recalibration using methods such as \textbf{conformal prediction}. These approaches could improve coverage while maintaining sufficiently precise intervals.

\section{Conclusion}

This study proposes an interpretable machine learning approach for predicting residential rents in Dakar. Starting from 1,654 raw listings, a rigorous preprocessing procedure made it possible to build a final corpus of 1,507 observations.

The XGBoost model optimized with Optuna achieves the best performance on the single split, with an R\textsuperscript{2} of 0.847, an MAE of 210,902 FCFA, and an RMSE of 324,195 FCFA.

The SHAP analysis furthermore shows that location constitutes one of the main determinants of rents in Dakar and reveals significant differences from the importance computed via XGBoost gain. These results underscore the value of combining several interpretability methods.

Finally, the intervals obtained via quantile regression show a coverage of 64.2\% against a target of 80\%. Recalibration therefore remains necessary before any operational use.

\subsection*{Implications for Housing Policy}

The rent differences observed between neighborhoods highlight the potential value of a predictive model for tracking market pressures and contributing to urban planning and affordable housing policies.

The main limitation remains the absence of data on actually negotiated rents. Establishing a structured system for collecting rental transaction data would improve the quality of available data and strengthen the tools available for analyzing Dakar's real estate market.

\subsection*{Limitations and Perspectives}

This study remains limited by the availability of online listings, the use of asking prices, the absence of precise geographic coordinates, and the uneven geographic coverage of the corpus. The prediction intervals also require recalibration.

Future work could incorporate more geospatial and temporal data, broaden data collection to different sources, and use actual transaction prices. Recalibration via conformal prediction could also improve the reliability of the prediction intervals.

\section*{Perspectives}

Several avenues could extend this work. Continuous data collection would make it possible to build a larger corpus and to better track market developments over time.

Adding precise geospatial variables such as distances to urban centers, transportation, beaches, and services would also help to better measure the effect of location.

Recalibrating the prediction intervals, notably using \textit{conformal prediction} methods, constitutes another priority for improving their reliability.

In the longer term, the creation of a national registry of rental transactions would make it possible to use actually negotiated rents and to develop models more representative of Senegal's real estate market.

\section*{Declaration}

\textbf{Funding}: this research received no external funding.

\textbf{Conflicts of interest}: no conflicts of interest declared.

\textbf{Data and code availability}: the dataset (\texttt{dataset.csv}, CC~BY~4.0 license) and the complete notebook documenting the entire pipeline are available upon request from the author.

\section*{References}

\begin{flushleft}
\setlength{\parindent}{0pt}
\setlength{\parskip}{0.6em}

ANSD: Population data \url{https://www.ansd.sn/Indicateur/donnees-de-population}

PressAfrik ANSD Study: Dakar, capital of renters \url{https://www.pressafrik.com/Etude-de-l-Ansd-Dakar-capitale-des-locataires_a219825.html}

Cybergeo Article on the real estate market in Dakar \url{https://journals.openedition.org/cybergeo/26146}

Abidoye, R.~B., \& Chan, A.~P.~C. (2018). Hedonic valuation of real estate properties in Nigeria. \textit{Journal of African Real Estate Research, 3}(1), 122--140. \url{https://pdfs.semanticscholar.org/9dd3/ef8baea1541e7438928532ebd498c89ffa88.pdf}

Chau, K.~W., \& Chin, T.~L. (2003). A critical review of literature on the hedonic price model. \textit{International Journal for Housing Science and Its Applications, 27}(2), 145--165. \url{https://ssrn.com/abstract=2073594}

Chen, T., \& Guestrin, C. (2016). XGBoost: A scalable tree boosting system. In \textit{Proceedings of the 22nd ACM SIGKDD International Conference on Knowledge Discovery and Data Mining} (pp.~785--794). ACM. \url{https://doi.org/10.1145/2939672.2939785}

Chica-Olmo, J., González-Morales, J.~G., \& Zafra-Gómez, J.~L. (2020). Effects of location on Airbnb apartment pricing in Málaga. \textit{Tourism Management, 77}, 103981. \url{https://doi.org/10.1016/j.tourman.2019.103981}

Irumba, R. (2015). An empirical examination of the effects of land tenure on housing values in Kampala, Uganda. \textit{International Journal of Housing Markets and Analysis, 8}(3), 359--374. \url{https://doi.org/10.1108/IJHMA-11-2014-0044}

Ke, G., Meng, Q., Finley, T., Wang, T., Chen, W., Ma, W., Ye, Q., \& Liu, T.-Y. (2017). LightGBM: A highly efficient gradient boosting decision tree. In \textit{Advances in Neural Information Processing Systems, 30}. \url{https://proceedings.neurips.cc/paper/2017/hash/6449f44a102fde848669bdd9eb6b76fa-Abstract.html}

Limsombunchai, V. (2004). House price prediction: Hedonic price model vs. artificial neural network. In \textit{Proceedings of the New Zealand Agricultural and Resource Economics Society Conference}.

Lundberg, S.~M., \& Lee, S.-I. (2017). A unified approach to interpreting model predictions. In \textit{Advances in Neural Information Processing Systems, 30}, 4765--4774. \url{https://papers.neurips.cc/paper/2017/hash/8a20a8621978632d76c43dfd28b67767-Abstract.html}

Lundberg, S.~M., Erion, G., Chen, H., DeGrave, A., Prutkin, J.~M., Nair, B., Katz, R., Himmelfarb, J., Bansal, N., \& Lee, S.-I. (2020). From local explanations to global understanding with explainable AI for trees. \textit{Nature Machine Intelligence, 2}(1), 56--67. \url{https://doi.org/10.1038/s42256-019-0138-9}

Pargent, F., Pfisterer, F., Thomas, J., \& Bischl, B. (2022). Regularized target encoding outperforms traditional methods in supervised machine learning with high cardinality features. \textit{Computational Statistics, 37}(5), 2671--2692. \url{https://doi.org/10.1007/s00180-022-01207-6}

Pérez-Rave, J.~I., Correa-Morales, J.~C., \& González-Echavarría, F. (2019). A machine learning approach to big data regression analysis of real estate prices for inferential and predictive purposes. \textit{Journal of Property Research, 36}(1), 59--96. \url{https://doi.org/10.1080/09599916.2019.1587489}

Rosen, S. (1974). Hedonic prices and implicit markets: Product differentiation in pure competition. \textit{Journal of Political Economy, 82}(1), 34--55. \url{https://doi.org/10.1086/260169}

Samb, R., Roy, U.~K., \& Suneja, M. (2025). Modeling the determinants of urban short-term rental prices in Gambia: A case of Greater Banjul Area. \textit{Research Square} [Preprint]. \url{https://doi.org/10.21203/rs.3.rs-8074344/v1}

Truong, Q., Nguyen, M., Dang, H., \& Mei, B. (2020). Housing price prediction via improved machine learning techniques. \textit{Procedia Computer Science, 174}, 433--442. \url{https://doi.org/10.1016/j.procs.2020.06.111}

\end{flushleft}

\end{document}